\documentclass[11pt]{article}
\usepackage{acl}
\usepackage{times}
\usepackage{latexsym}
\usepackage[T1]{fontenc}
\usepackage{CJKutf8}
\usepackage{inconsolata}
\usepackage{graphicx}
\usepackage{tabularx}
\usepackage{booktabs}
\usepackage{caption}
\usepackage{float}
\usepackage{listings}
\usepackage{colortbl}
\usepackage[table]{xcolor}
\usepackage{subcaption}
\usepackage{enumitem}
\setlist{nosep}

\title{A Comparative Study of LLM Prompting and Fine-Tuning for Cross-genre Authorship Attribution on Chinese Lyrics}

\author{Yuxin Li, Lorraine Xu \and Meng Fan Wang \\
        Columbia University \\
        \texttt{\{yl5574, yx2950, mw3751\}@columbia.edu}}

\begin{document}
\maketitle

\begin{abstract}
We propose a novel study on \textit{authorship attribution} for Chinese lyrics—a domain where clean, public datasets are sorely lacking. Our contributions are twofold: (1) we create a new, balanced dataset of Chinese lyrics spanning multiple genres, and (2) we develop and fine-tune a domain-specific model, comparing its performance against zero-shot inference using the DeepSeek LLM.

We test two central hypotheses. First, we hypothesize that a fine-tuned model will outperform a zero-shot LLM baseline. Second, we hypothesize that performance is \emph{genre-dependent}. Our experiments strongly confirm Hypothesis\,2: structured genres (e.g., \emph{\begin{CJK}{UTF8}{gbsn} 民俗与传统 \end{CJK}}, Folklore \& Tradition) yield significantly higher attribution accuracy than more abstract genres (e.g., \emph{\begin{CJK}{UTF8}{gbsn} 爱与浪漫 \end{CJK}}, Love \& Romance). Hypothesis\,1 receives only partial support: fine-tuning improves robustness and generalization in Test1 (real-world data and difficult genres), but offers limited or ambiguous gains in Test2, a smaller, synthetically-augmented set. We show that the design limitations of Test2 (e.g., label imbalance, shallow lexical differences, and narrow genre sampling) can obscure the true effectiveness of fine-tuning.

Our work establishes the first benchmark for cross-genre Chinese lyric attribution, highlights the importance of genre-sensitive evaluation, and provides a public dataset and analytical framework for future research. We conclude with recommendations: enlarge and diversify test sets, reduce reliance on token-level data augmentation, balance author representation across genres, and investigate domain-adaptive pretraining as a pathway for improved attribution performance.
\end{abstract}

\section{Introduction and Motivation}
Our work focuses on authorship attribution for Chinese lyrics. To the best of our knowledge, there is a lack of prior research and clean, publicly available datasets in this domain. The goal of this project is to develop and fine-tune a domain-specific model for authorship attribution on cross-genre Chinese lyrics, while also constructing a new, balanced dataset of Chinese lyrics spanning across multiple genres. In addition, the project aims to evaluate the effectiveness of this fine-tuned model by comparing its performance with existing large language models (LLMs), i.e. DeepSeek. Through this work, the project seeks to fill a gap in authorship research on Chinese lyrics and contribute valuable resources and benchmarks for future studies.

\section{Research Question and Hypotheses}

Our research question is how does a fine-tuned, domain-specific model compare to general-purpose large language models in authorship attribution of cross-genre Chinese lyrics, and how does attribution performance vary across lyrical genres?

\textbf{Hypothesis 1}: A fine-tuned domain-specific model will outperform zero-shot in-context learning with general-purpose LLMs (i.e. DeepSeek) in authorship attribution of cross-genre Chinese lyrics.

\textbf{Hypothesis 2}: Authorship attribution performance will vary between genres, with certain genres yielding higher accuracy due to more distinct stylistic features.

\section{Related Work}

To address our research question, we need to define what is meant by genres of lyrics. In fact, the genre classification of lyrics has always been a difficult problem in both music and literature. In this paper, we focus on the topics and themes expressed in the lyrics of the songs. Rather than adopting conventional genre labels from music information retrieval, such as pop, country, or folk, as commonly used in lyric genre classification \citep{tsaptsinos:2017, mayer:2008}, we draw inspiration from the approach in \citet{czedik-Eysenberg:2019}, which categorizes lyrical content based on thematic topics found in metal music. Taking into account the lyrics in our dataset and the cultural differences between China and the West, we adapt the Czedik-Eysenberg genre definitions. For example, we replace religiously related tags with the category 'Tradition / Folklore' and propose the categories, shown in Table~\ref{tab:lyric-genres}, as a baseline for our study.

\begin{table}[t]
  \centering
  \begin{tabularx}{\linewidth}{l l}
    \hline
    \textbf{\begin{CJK}{UTF8}{gbsn} 中文标签 \end{CJK}} & \textbf{English Tag} \\
    \hline
    \begin{CJK}{UTF8}{gbsn} 民间与传统 \end{CJK} & Folklore \& Tradition \\
    \begin{CJK}{UTF8}{gbsn} 爱与浪漫 \end{CJK} & Love \& Romance \\
    \begin{CJK}{UTF8}{gbsn} 生活与反思 \end{CJK} & Life \& Reflection \\
    \begin{CJK}{UTF8}{gbsn} 社会与现实 \end{CJK} & Society \& Reality \\
    \begin{CJK}{UTF8}{gbsn} 风景与旅程 \end{CJK} & Landscape \& Journey \\
    \hline
  \end{tabularx}
  \caption{Theme-based lyric genre categories used in our study. \\ Adapted to reflect cultural context in Chinese lyrics.}
  \label{tab:lyric-genres}
\end{table}

After defining these categories, we follow a human-AI collaboration approach inspired by \citet{Wang:2023}, in which we use DeepSeek to automatically assign genre tags to the lyrics and then manually verify the precision of the labels.

We draw on the work of \citet{Sahin:2021}, which suggests that token-level augmentation can help mitigate data sparsity in natural language processing learning. Following this insight, we augment positive samples in our sparse dataset by inserting random tokens to enhance representation diversity.

To ensure our model learns true stylistic features rather than relying on superficial topic-author correlations, we adopt the concept of “hard positives” and “hard negatives” from \citet{fincke:2024} in our dataset design. Specifically, we treat lyrics written by the same author but in different genres as hard positives, encouraging the model to capture consistent stylistic traits across topics. Conversely, we select lyrics written by different authors within similar thematic domains as hard negatives, making the classification task more challenging and reducing the risk of topic-based shortcuts. We use Chinese-RoBERTa \footnote{\url{https://huggingface.co/hfl/chinese-roberta-wwm-ext}} \citep{xu:2021} as our base model for fine-tuning, as it was pretrained and further optimized for Chinese text classification tasks.

Inspired by the findings of \citet{huang:2024} that large language models "excel at identifying authorship without the need for domain-specific fine-tuning," we apply Linguistically Informed Prompting to enable zero-shot in-context learning using DeepSeek. We then conduct a comparative analysis between the performance of this LLM-based approach and our fine-tuned Chinese-RoBERTa model.

To conduct the comparative study, we follow the evaluation guidelines provided by PAN\footnote{\url{https://pan.webis.de/clef23/pan23-web/author-identification.html\#evaluation}}. Our baseline setup involves presenting the model with a pair of lyrics and asking it to determine whether they were written by the same lyricist.

\section{Technical Approach}

\subsection{Raw Dataset}
\subsubsection{Raw Dataset Information}
\label{sec:datasetinfo}
We collect 1055 Chinese songs, all of which were released after the year 2000 and contain more than 80\% Chinese text. Each entry in the dataset includes the following attributes: title, lyricist, genre, length, and lyrics. 

We divide the 1055 songs into 2 groups to serve our evaluation purpose, which is explained in Section~\ref{sec:trainandtest}. The first 1000 songs serve for Training and Test1, while the other 55 songs are found at a later stage as the Test2 dataset, among which none of the lyricists appear in the training or Test1 datasets. Each lyricist in the Training and Test1 datasets is represented by 4 to 30 songs, while this requirement is not strictly applied to Test2 due to time constraints within the scope of our project. The details of splitting the datasets will be explained in the end of the Section~\ref{sec:test2rawdataset}.

The attributes “Title,” “Lyricist,” and “Raw Lyrics” are manually collected from sources such as Baidu Baike\footnote{\url{https://baike.baidu.com/}}, Gequbao, and the QQ Music mobile app\footnote{\url{https://y.qq.com/}}. The “Length” attribute, representing the number of lines in each song, is computed using a Python script. The “Genre” field is assigned based on our baseline labeling method using Deepseek-Lllama-70B\footnote{\url{https://www.together.ai/models/deepseek-r1-distilled-llama-70b-free}}.

\subsubsection{Raw Lyrics Post-process}
\begin{figure}[t]
  \includegraphics[width=\columnwidth]{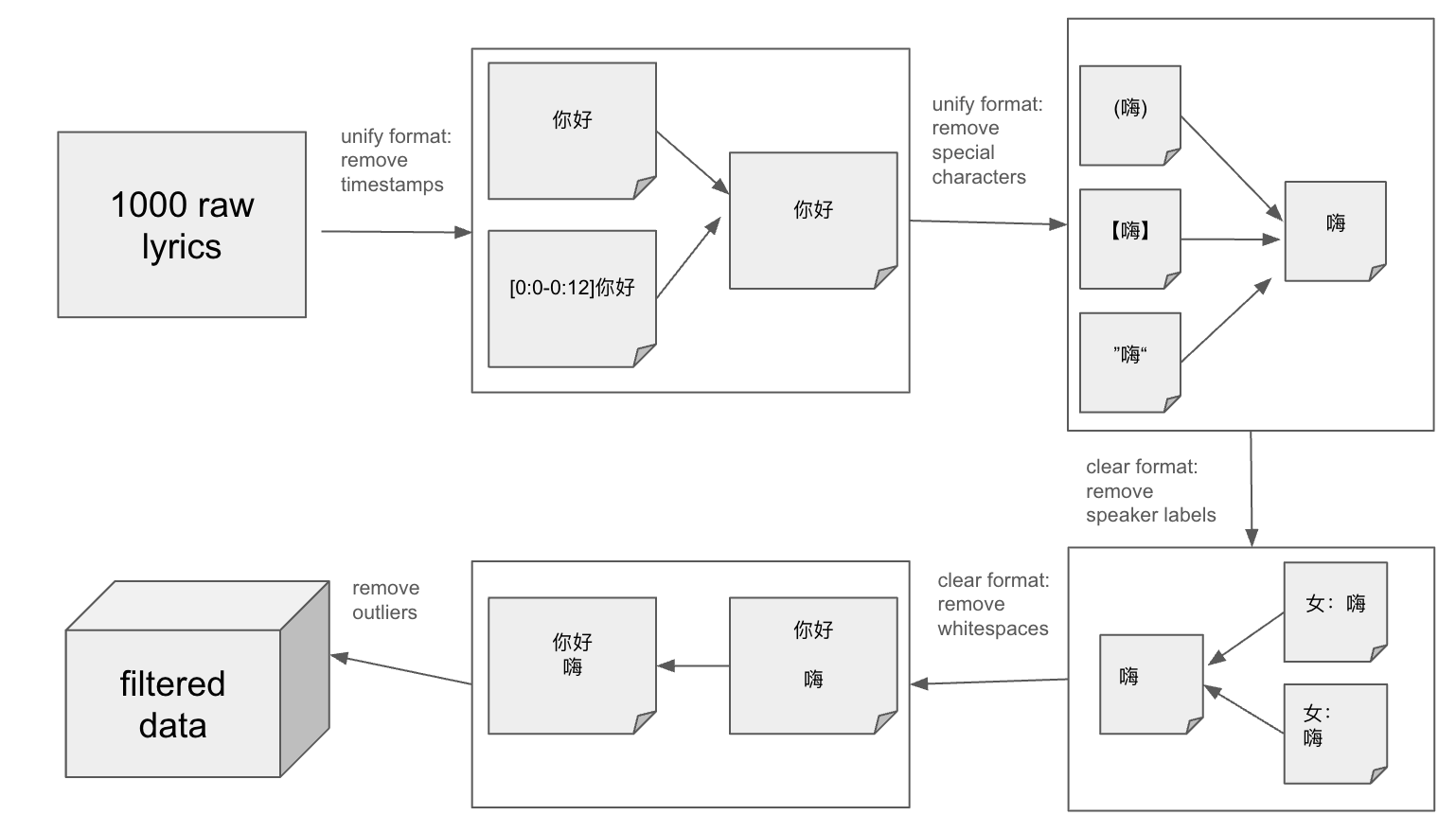}
  \caption{Workflow for cleaning and filtering data.}
  \label{fig:datafiltering}
\end{figure}

We apply a data-filtering pipeline to standardize the content of raw lyrics. We remove timestamps, speaker labels, and special characters such as brackets or parentheses that do not contribute to the actual lyrics. We also strip away any extraneous text preceding speaker labels to ensure only the lyrical content remained. After this step, we use a Python script to extract the “Length” information. Of the first 1,000 songs, the lengths range from 16 to 119 lines, with a median of 43.0 and a mean of 45.696. To prevent LLMs from relying on song lengths to make authorship attribution, we generate a box plot to visualize the distribution of song lengths, identifying and removing 31 outliers based on Tukey's Fences method\footnote{\url{https://aakinshin.net/posts/tukey-outlier-probability/}}. The complete data-filtering workflow is illustrated in Figure~\ref{fig:datafiltering}. Since the Test2 dataset is created separately from Training and Test1, instead of removing outliers in Test2 using its own statistics, we control the length range of Test2 to be within the lower (10.5) and upper (78.5) bound of the first 1000 songs. The song length statistics and distributions are shown in Appendix~\ref{sec:appendixA1}.

\subsubsection{Genre Generation}
We apply a genre-generation pipeline with Deepseek-Lllama-70B to assign genres to lyrics. We sample 10 out of 1000 songs and test two settings: English prompt with Chinese lyrics for English output, and Chinese prompt with Chinese lyrics for Chinese output. The prompts include genre concepts, expected output format, and the possibility of multiple genres if necessary. In early attempts, the model generate multiple tags for all samples, which defeats the purpose of classifying lyrics based on genres. We revise the prompt to ensure multiple genres are only assigned when equally likely, with a preference for a single genre. After the model generates the results for both settings, we manually evaluate the genre classification accuracy and find the Chinese prompt to be more accurate (Appendix~\ref{sec:genre-prompt-comparison}), so we apply the Chinese prompt to the full dataset. We also complete a second round of human evaluation once the results for the full dataset were generated. The genre-generation workflow is shown in Figure~\ref{fig:genregeneration}. The exact Chinese prompt used is provided in Appendix~\ref{sec:genre-chinese-prompt}, and the English prompt is shown below:

\noindent
\begin{minipage}{\linewidth}
\begin{lstlisting}[basicstyle=\ttfamily\tiny, frame=single, breaklines=true, breakatwhitespace=true, numbers=none]
Given the following genre concepts:
1. Love & Romance: Romance, heartbreak, longing, missing someone
2. Life & Reflection: Growth, regret, personal lessons, contemplation
3. Society & Reality: Urban struggles, inequality, class, political tone
4. Landscape & Journey: Nature, travel, scenery, wandering, solitude
5. Folklore & Tradition: Legends, cultural icons, regional storytelling, historical motifs
Please classify the genre of the following lyrics:
{lyrics}
If the model thinks multiple genres are equally likely to be the genre of these lyrics, it can generate multiple genres. However, in most cases, only one genre should be provided. The output should follow the format:
Genres: [Genre1]
Where Genre1 is the genre label (e.g., Love & Romance, Life & Reflection). If multiple genres are equally likely, they should be listed inside the square brackets and separated by commas, e.g., Genres: [Love & Romance, Life & Reflection]. But again, most of the time, there should only be one genre listed inside the brackets.
\end{lstlisting}
\end{minipage}

\begin{figure}[t]
  \centering
  \includegraphics[width=0.60\columnwidth]{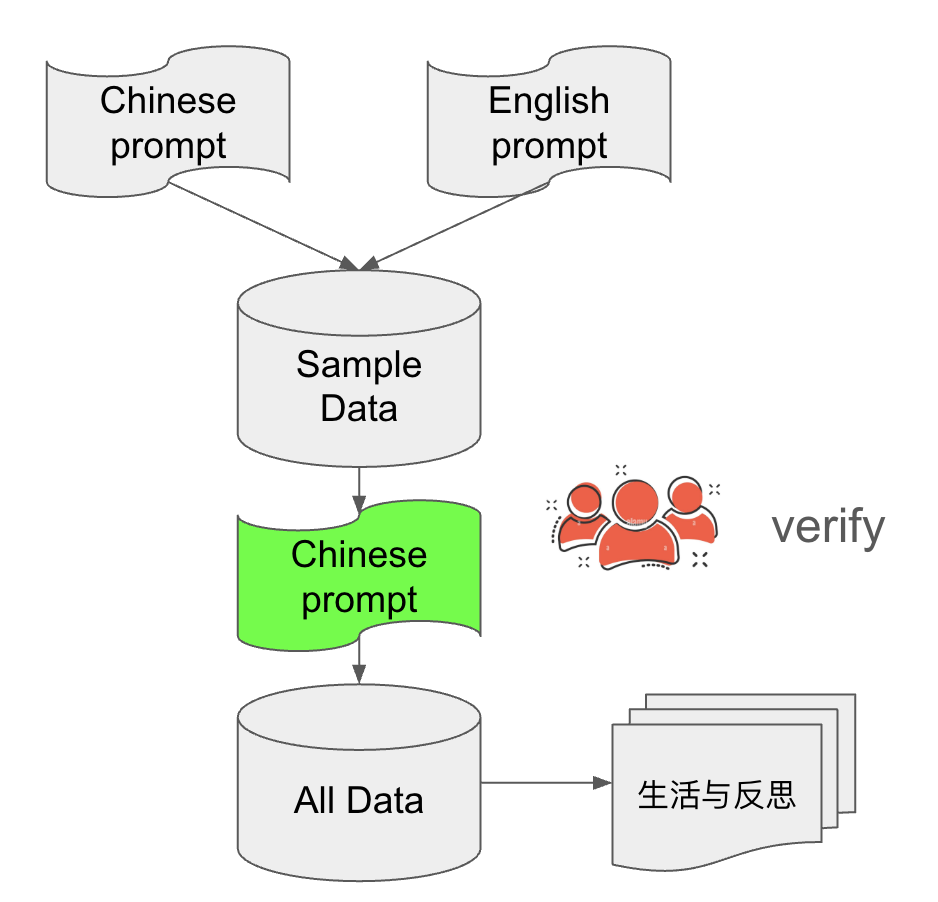}
  \caption{Workflow for generating genres.}
  \label{fig:genregeneration}
\end{figure}

\subsubsection{Build the Second, Smaller Dataset}
\label{sec:test2rawdataset}
After generating genres for the first 1000 songs, we count the numbers of unique lyricists in each of the five genres and plot them with a histogram (Lyricists with songs in multiple genres are counted as unique authors in each of the genres their songs appear in). Based on the distribution of unique authors across genres for the Training and Test1 datasets, we purposefully look for songs with a similar genre distribution to construct the Test2 dataset. This approach ensures the consistency between our training and the 2 testing datasets. After a careful analysis for the distribution, we downscale our plan to find a minimum of 55 songs for the second dataset. 

However, we soon observe that the sparsity of the raw Test2 data poses significant challenges in constructing pairwise testing examples, as detailed in Section~\ref{sec:trainandtest}, ultimately leading to irregularities in model evaluation. To address this, and inspired by \citet{Sahin:2021}, we introduce synthetic data by injecting tokens such as [SYN0] into the raw dataset. This approach allows us to balance the dataset while maintaining a distribution similar to the original.

\begin{figure*}[t]
    \centering
    \begin{minipage}{0.3\textwidth}
        \centering
        \includegraphics[width=\linewidth]{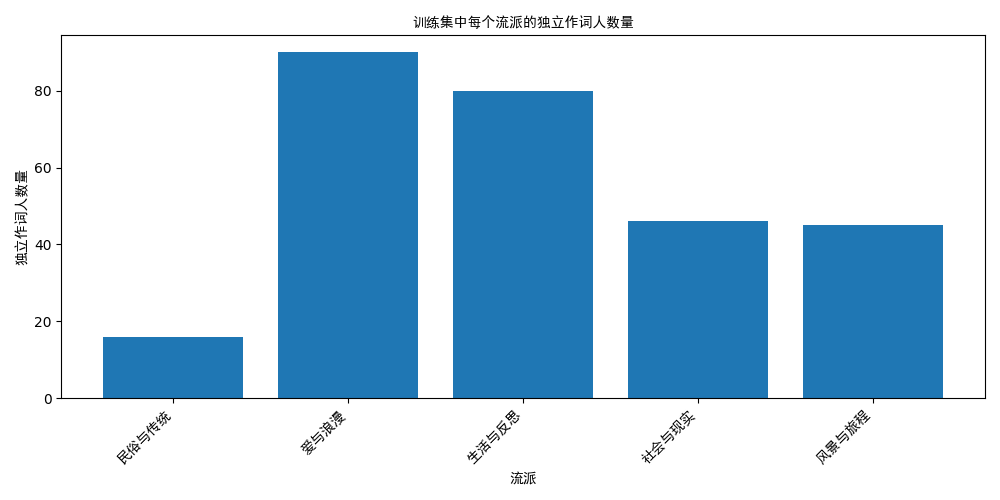}
        \subcaption{Training Data}
        \label{fig:train_data_hist}
    \end{minipage}
    \hfill
    \begin{minipage}{0.3\textwidth}
        \centering
        \includegraphics[width=\linewidth]{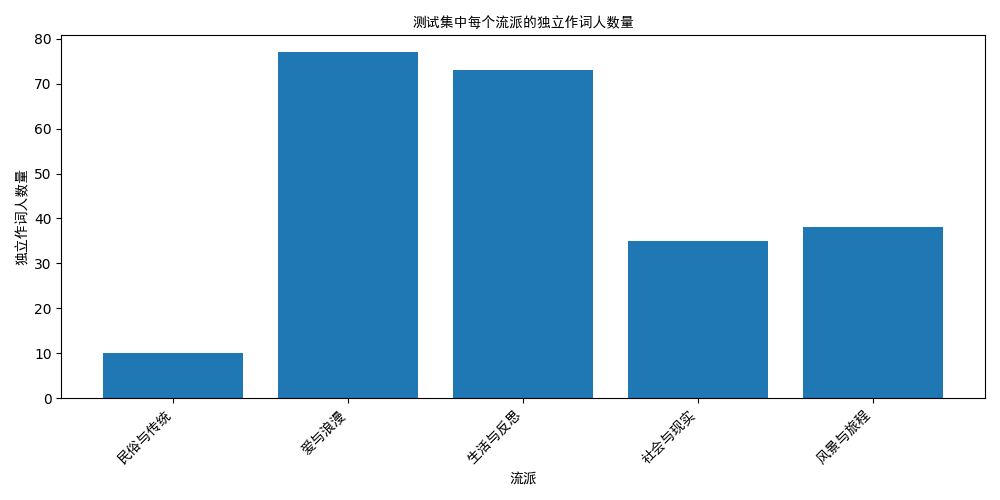}
        \subcaption{Test1 Data}
        \label{fig:test_data_1_hist}
    \end{minipage}
    \hfill
    \begin{minipage}{0.3\textwidth}
        \centering
        \includegraphics[width=\linewidth]{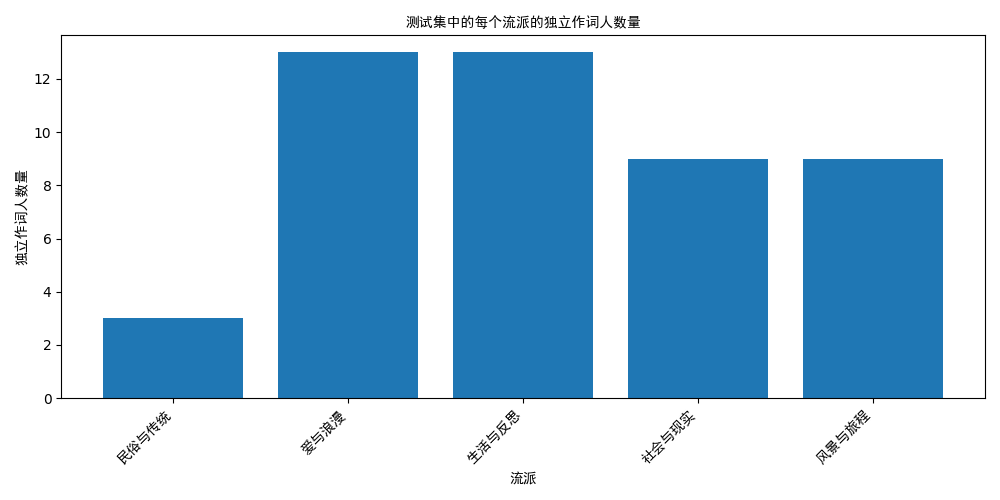}
        \subcaption{Test2 Data}
        \label{fig:test_data_2_hist}
    \end{minipage}
    \caption{Histograms of unique lyricists per genre for the training and test datasets.}
    \label{fig:uniqueauthordistribution}
\end{figure*}

The distributions of unique lyricists across genres for training, Test1, and Test2 are shown in Figure~\ref{fig:uniqueauthordistribution}.

\subsection{Train and Test Data}
\label{sec:trainandtest}
As mentioned in Section~\ref{sec:datasetinfo}. We build 3 splits, Train, Test1, and Test2, where Train and Test1 are created from the 1,000-song set with a split ratio of 80:20, and Test2 contains lyrics from previously unseen authors in the training. Both Test1 and Test2 have no overlap pairs from Training. In this way, we hope to accurately evaluate the model's performance and test its generalization ability. 

We construct our training and evaluation datasets using pairwise comparisons. For both per-genre and cross-genre settings, we generate pairs consisting of lyrics written by the same author (label = 1) and by different authors (label = 0). To ensure balanced evaluation, we carefully curate the dataset so that all five genres include both label types and both settings. Also, the proportions of overall per-genre vs. cross-genre pairs, as well as same-author vs. different-author pairs, are kept relatively consistent across the Train, Test1, and Test2 splits. The detailed statistics per genre and per setting in each split are provided in Appendix~\ref{sec:appendixA2}, Appendix~\ref{sec:appendixA3}, and Appendix~\ref{sec:appendixA4}.

Our construction results in the following number of pairwise examples for each split:
\begin{itemize}[noitemsep, topsep=0pt]
  \item \textbf{Train Set} – 965 pairs
  \item \textbf{Test1 Set} – 244 pairs
  \item \textbf{Test2 Set} – 68 pairs
\end{itemize}

\subsection{Metrics}
\label{sec:metrics}
We report standard evaluation metrics, including F1-scores, accuracy, precision, and recall. Specifically, we use F1-Micro, F1-Macro, and F1-Weighted scores to examine potential label imbalances and provide a more accurate evaluation. Additionally, to analyze performance across different genres, we compute per-class metrics for a more fine-grained assessment.

\subsection{Zero Shot}
We evaluate our dataset using Deepseek-Lllama-70B in a zero-shot setting. To prompt the model for authorship attribution, we design a linguistically informed prompt that encourages the model to focus on stylistic differences rather than topical content. Notably, since prompting in Chinese outperforms English when generating genres for Chinese lyrics, we adopt a Chinese prompt for the authorship attribution task as well.

The exact Chinese prompt used is provided in Appendix~\ref{sec:chinese-prompt}, and the English translation is shown below:

\noindent
\begin{minipage}{\linewidth}
\begin{lstlisting}[basicstyle=\ttfamily\footnotesize, frame=single, breaklines=true, breakatwhitespace=true, numbers=none]
Determine whether the two input texts were written by the same author.
Analyze the writing style of the texts while ignoring differences in topic and content.
The reasoning should be based on linguistic features such as verb usage, punctuation,
rare vocabulary, affixes, humor, sarcasm, typos, and spelling variations.
The output should follow this format: 0 or 1 (0 indicates different authors, 1 indicates the same author).
\end{lstlisting}
\end{minipage}

\subsection{Fine Tune}
We fine-tune the pre-trained Chinese RoBERTa base model (hfl/chinese-roberta-wwm-ext) using contrastive learning. Specifically, we employ a custom ContrastiveLoss from Sentence Transformer, which optimizes the model to increase cosine similarity for positive pairs (label = 1) and decrease it for negative pairs (label = 0), effectively separating dissimilar pairs in the embedding space.

Our dataset is split 80/20 into training and validation sets. We train the model for 2 epochs using the Adam optimizer with a learning rate of 2e-5, a batch size of 16, and a maximum sequence length of 128. A warm-up phase of 100 steps is used to stabilize training dynamics early on.

For evaluation, we use EmbeddingSimilarityEvaluator along with a custom callback that computes macro F1 scores across a range of cosine similarity thresholds. The callback updates the best-performing threshold every 50 steps based on validation results, enabling dynamic tracking of classification performance\footnote{Our training setup is inspired by \url{https://github.com/UKPLab/sentence-transformers/blob/master/examples/sentence_transformer/training/nli/training_nli_v2.py} and the customized contrastiveloss follows \url{https://github.com/UKPLab/sentence-transformers/blob/master/sentence_transformers/losses/ContrastiveLoss.py}}.

\section{Results and Error Analysis}
\subsection{Results}
Zero-shot results on the Test1 and Test2 splits are summarized in Tables~\ref{tab:genre-metrics-test1} and~\ref{tab:genre-metrics-test2}, respectively. Tables~\ref{tab:genre-metrics-test3} and~\ref{tab:genre-metrics-test4} show the results for finetuning. For each table, the highest score in each evaluation metric column is highlighted in gray to emphasize peak performance across genres and modes.

\begin{table*}[t]
\centering
\begin{minipage}[t]{0.49\textwidth}
\caption{Evaluation metrics per genre and mode for Test1 (finetune). The genres are shown in the sequence of Folklore \& Tradition, Love \& Romance, Life \& Reflection, Society \& Reality, Landscape \& Journey}
\label{tab:genre-metrics-test1}
\scriptsize
\resizebox{\linewidth}{!}{
\begin{tabular}{llcccccc}
\toprule
Genre & Mode & Accuracy & F1 Micro & F1 Weighted & F1 Macro & Recall & Precision \\
\midrule
Folklore \& Tradition & per-genre    & 0.7500 & 0.7500 & 0.7333 & 0.7333 & 0.7500 & 0.8333 \\
Folklore \& Tradition & cross-genre  & \cellcolor{gray!30}0.8889 & \cellcolor{gray!30}0.8889 & \cellcolor{gray!30}0.8821 & \cellcolor{gray!30}0.8615 & \cellcolor{gray!30}0.8333 & \cellcolor{gray!30}0.9286 \\
Love \& Romance & per-genre    & 0.5714 & 0.5714 & 0.5510 & 0.5625 & 0.6185 & 0.6469 \\
Love \& Romance & cross-genre  & 0.5325 & 0.5325 & 0.4815 & 0.4273 & 0.4603 & 0.4365 \\
Life \& Reflection & per-genre    & 0.6429 & 0.6429 & 0.6362 & 0.6393 & 0.6563 & 0.6689 \\
Life \& Reflection & cross-genre  & 0.4925 & 0.4925 & 0.4552 & 0.4482 & 0.4808 & 0.4737 \\
Society \& Reality & per-genre    & 0.5385 & 0.5385 & 0.5064 & 0.4583 & 0.4750 & 0.4667 \\
Society \& Reality & cross-genre  & 0.4359 & 0.4359 & 0.3650 & 0.3871 & 0.4799 & 0.4621 \\
Landscape \& Journey & per-genre    & 0.7778 & 0.7778 & 0.7579 & 0.6786 & 0.6615 & 0.7333 \\
Landscape \& Journey & cross-genre  & 0.5366 & 0.5366 & 0.4868 & 0.5073 & 0.5956 & 0.6639 \\
\midrule
\textbf{Overall} & per-genre    & \cellcolor{gray!30}0.5827 & \cellcolor{gray!30}0.5827 & \cellcolor{gray!30}0.5579 & \cellcolor{gray!30}0.5637 & \cellcolor{gray!30}0.5988 & \cellcolor{gray!30}0.6281 \\
\textbf{Overall} & cross-genre  & 0.5238 & 0.5238 & 0.4727 & 0.4563 & 0.4973 & 0.4956 \\
\bottomrule
\end{tabular}
}
\end{minipage}
\hfill
\begin{minipage}[t]{0.49\textwidth}
\caption{Evaluation metrics per genre and mode for Test1 (zero-shot). The genres are shown in the sequence of Folklore \& Tradition, Love \& Romance, Life \& Reflection, Society \& Reality, Landscape \& Journey}
\label{tab:genre-metrics-test3}
\scriptsize
\resizebox{\linewidth}{!}{
\begin{tabular}{llcccccc}
\toprule
Genre & Mode & Accuracy & F1 Micro & F1 Weighted & F1 Macro & Recall & Precision \\
\midrule
Folklore \& Tradition & cross-genre  & 0.4167 & 0.4167 & 0.3259 & 0.3778 & 1.0000 & 0.3636 \\
Folklore \& Tradition & per-genre    & \cellcolor{gray!30}1.0000 & \cellcolor{gray!30}1.0000 & \cellcolor{gray!30}1.0000 & \cellcolor{gray!30}1.0000 & \cellcolor{gray!30}1.0000 & \cellcolor{gray!30}1.0000 \\
Love \& Romance & cross-genre  & 0.4130 & 0.4130 & 0.2920 & 0.3432 & 0.9714 & 0.3908 \\
Love \& Romance  & per-genre    & 0.6506 & 0.6506 & 0.5247 & 0.4249 & \cellcolor{gray!30}1.0000 & 0.6463 \\
Life \& Reflection & cross-genre  & 0.5376 & 0.5376 & 0.3976 & 0.3879 & \cellcolor{gray!30}1.0000 & 0.5275 \\
Life \& Reflection & per-genre    & 0.5455 & 0.5455 & 0.4271 & 0.4271 & \cellcolor{gray!30}1.0000 & 0.5238 \\
Society \& Reality & cross-genre  & 0.5854 & 0.5854 & 0.4711 & 0.4177 & 0.9583 & 0.5897 \\
Society \& Reality & per-genre    & 0.2727 & 0.2727 & 0.1169 & 0.2143 & \cellcolor{gray!30}1.0000 & 0.2727 \\
Landscape \& Journey & cross-genre  & 0.6275 & 0.6275 & 0.5436 & 0.5215 & \cellcolor{gray!30}1.0000 & 0.5957 \\
Landscape \& Journey & per-genre    & 0.1250 & 0.1250 & 0.0278 & 0.1111 & \cellcolor{gray!30}1.0000 & 0.1250 \\
\midrule
\textbf{Overall} & cross-genre  & 0.5156 & 0.5156 & 0.3943 & \cellcolor{gray!30}0.4041 & 0.9856 & 0.4982 \\
\textbf{Overall} & per-genre    & \cellcolor{gray!30}0.5646 & \cellcolor{gray!30}0.5646 & \cellcolor{gray!30}0.4278 & 0.4000 & \cellcolor{gray!30}1.0000 & \cellcolor{gray!30}0.5556 \\
\bottomrule
\end{tabular}
}
\end{minipage}
\end{table*}

\begin{table*}[t]
\centering
\begin{minipage}[t]{0.49\textwidth}
\caption{Evaluation metrics per genre and mode for Test2 (finetune). The genres are shown in the sequence of Folklore \& Tradition, Love \& Romance, Life \& Reflection, Society \& Reality, Landscape \& Journey}
\label{tab:genre-metrics-test4}
\scriptsize
\resizebox{\linewidth}{!}{
\begin{tabular}{llcccccc}
\toprule
Genre & Mode & Accuracy & F1 Micro & F1 Weighted & F1 Macro & Recall & Precision \\
\midrule
Folklore \& Tradition & cross-genre  & 0.6250 & 0.6250 & 0.4808 & 0.3846 & \cellcolor{gray!30}1.0000 & 0.6250 \\
Folklore \& Tradition & per-genre    & \cellcolor{gray!30}1.0000 & \cellcolor{gray!30}1.0000 & \cellcolor{gray!30}1.0000 & \cellcolor{gray!30}1.0000 & \cellcolor{gray!30}1.0000 & \cellcolor{gray!30}1.0000 \\
Love \& Romance & cross-genre  & 0.3750 & 0.3750 & 0.2045 & 0.2727 & \cellcolor{gray!30}1.0000 & 0.3750 \\
Love \& Romance   & per-genre    & 0.1250 & 0.1250 & 0.0278 & 0.1111 & \cellcolor{gray!30}1.0000 & 0.1250 \\
Life \& Reflection & cross-genre  & 0.3636 & 0.3636 & 0.1939 & 0.2667 & \cellcolor{gray!30}1.0000 & 0.3636 \\
Life \& Reflection & per-genre    & 0.4615 & 0.4615 & 0.2915 & 0.3158 & \cellcolor{gray!30}1.0000 & 0.4615 \\
Society \& Reality & cross-genre  & 0.5000 & 0.5000 & 0.3333 & 0.3333 & \cellcolor{gray!30}1.0000 & 0.5000 \\
Society \& Reality & per-genre    & 0.5000 & 0.5000 & 0.3333 & 0.3333 & \cellcolor{gray!30}1.0000 & 0.5000 \\
Landscape \& Journey & cross-genre  & 0.3125 & 0.3125 & 0.1488 & 0.2381 & \cellcolor{gray!30}1.0000 & 0.3125 \\
Landscape \& Journey & per-genre    & 0.5000 & 0.5000 & 0.3333 & 0.3333 & \cellcolor{gray!30}1.0000 & 0.5000 \\
\midrule
\textbf{Overall} & cross-genre  & 0.4118 & 0.4118 & 0.2402 & 0.2917 & \cellcolor{gray!30}1.0000 & 0.4118 \\
\textbf{Overall} & per-genre    & 0.4118 & 0.4118 & 0.2402 & 0.2917 & \cellcolor{gray!30}1.0000 & 0.4118 \\
\bottomrule
\end{tabular}
}
\end{minipage}
\hfill
\begin{minipage}[t]{0.49\textwidth}
\caption{Evaluation metrics per genre and mode for Test2 (zero-shot). The genres are shown in the sequence of Folklore \& Tradition, Love \& Romance, Life \& Reflection, Society \& Reality, Landscape \& Journey}
\label{tab:genre-metrics-test2}
\scriptsize
\resizebox{\linewidth}{!}{
\begin{tabular}{llcccccc}
\toprule
Genre & Mode & Accuracy & F1 Micro & F1 Weighted & F1 Macro & Recall & Precision \\
\midrule
Folklore \& Tradition & per-genre    & \cellcolor{gray!30}1.0000 & \cellcolor{gray!30}1.0000 & \cellcolor{gray!30}1.0000 & \cellcolor{gray!30}1.0000 & \cellcolor{gray!30}1.0000 & \cellcolor{gray!30}1.0000 \\
Folklore \& Tradition & cross-genre  & 0.5000 & 0.5000 & 0.4333 & 0.4667 & 0.6000 & 0.7143 \\
Love \& Romance & per-genre    & 0.8750 & 0.8750 & 0.8910 & 0.7949 & 0.9286 & 0.7500 \\
Love \& Romance & cross-genre  & 0.5000 & 0.5000 & 0.5000 & 0.4667 & 0.4667 & 0.4667 \\
Life \& Reflection & per-genre    & 0.7692 & 0.7692 & 0.7692 & 0.7692 & 0.7738 & 0.7738 \\
Life \& Reflection & cross-genre  & 0.6818 & 0.6818 & 0.6767 & 0.6460 & 0.6429 & 0.6524 \\
Society \& Reality & per-genre    & 0.5000 & 0.5000 & 0.3333 & 0.3333 & 0.5000 & 0.2500 \\
Society \& Reality & cross-genre  & 0.5714 & 0.5714 & 0.5333 & 0.5333 & 0.5714 & 0.6061 \\
Landscape \& Journey & per-genre    & \cellcolor{gray!30}1.0000 & \cellcolor{gray!30}1.0000 & \cellcolor{gray!30}1.0000 & \cellcolor{gray!30}1.0000 & \cellcolor{gray!30}1.0000 & \cellcolor{gray!30}1.0000 \\
Landscape \& Journey & cross-genre  & 0.8571 & 0.8571 & 0.9231 & 0.4615 & 0.4286 & 0.5000 \\
\midrule
\textbf{Overall} & per-genre    & \cellcolor{gray!30}0.8235 & \cellcolor{gray!30}0.8235 & \cellcolor{gray!30}0.8248 & \cellcolor{gray!30}0.8229 & \cellcolor{gray!30}0.8393 & \cellcolor{gray!30}0.8299 \\
\textbf{Overall} & cross-genre  & 0.5882 & 0.5882 & 0.5589 & 0.5296 & 0.5429 & 0.5577 \\
\bottomrule
\end{tabular}
}
\end{minipage}
\end{table*}

We evaluate our results by reflecting on our hypotheses.

\textbf{Hypothesis 1} posits that a fine-tuned domain-specific model will outperform zero-shot in-context learning with general-purpose LLMs, i.e. DeepSeek, in cross-genre authorship attribution.

Overall, the comparison supports the first hypothesis. The finetuned model generally achieves higher performance across most metrics, especially in structured genres like Folklore \& Tradition and to a lesser degree in Landscape \& Journey. In Test1, finetuning improves overall F1 Macro (from 0.4563 zero-shot cross-genre to 0.4041 finetuned, and from 0.5637 zero-shot per-genre to 0.4000 finetuned). Even for genres that do not have a significant improvement, we observe that the finetuned model performs comparably similar to the LLM, which already shows the power of finetuning pretrained model given that we only use 900+ data for training.

In Test2, the overall metrics (F1 Macro: 0.2917) for both modes are lower than in the zero-shot version (0.5296 for cross-genre and 0.8229 for per-genre). Since Test2 is specifically constructed with unseen authors to evaluate out-of-distribution generalizability, the lower performance of our finetuned model is expected. Compared to large language models (LLMs), which are trained on vast corpora and may have already encountered similar or even overlapping content, our model exhibits more limited generalization capacity. Nonetheless, the observed F1 Macro score of 0.2917 demonstrates that the model retains some ability to generalize beyond the training authors, which is an encouraging baseline given the data and resource constraints of our finetuning setup. Also, we hope our model might perform better on real life data rather than our synthetic ones.

Thus, finetuning appears to offer meaningful gains in robustness and generalizability. However, its benefits are best observed in real test sets like Test1, while synthetic sets like Test2 may obscure its full effect. And hypothesis 1 is partially supported.

\textbf{Hypothesis 2} states that authorship attribution performance will vary across genres, depending on the presence of distinct stylistic features.

Folklore \& Tradition consistently performs well across all settings. In the zero-shot setup, it achieves perfect or near-perfect scores in per-genre mode for both Test1 and Test2. After finetuning, this strength persists—maintaining a perfect score (1.0000) in per-genre mode, while also showing improved generalization in cross-genre mode in Test2 (Accuracy: 0.6250, F1 Macro: 0.3846). These results suggest that this genre is highly stylized, with strong intra-author coherence, and that finetuning enhances generalization without sacrificing performance in structured settings.

Love \& Romance, by contrast, exhibits inconsistent and generally weaker performance. In the zero-shot condition, it achieves moderate results in Test1 per-genre (F1 Macro: 0.5625) but degrades in Test2. Finetuning brings small improvements in Test1 (F1 Weighted: 0.5247), yet in Test2 per-genre mode, the model collapses (F1 Macro: 0.1111). These patterns likely stem from the emotional and generic language prevalent in the genre, which introduces semantic overlap across authors and reduces the distinctiveness needed for attribution. Finetuning offers limited benefit here.

Life \& Reflection shows relatively stable and moderate performance across settings. In the zero-shot setup, it performs particularly well in Test2 cross-genre (F1 Macro: 0.6460), and maintains similar mid-range results after finetuning, with F1 Macro scores generally ranging from 0.31 to 0.43. Slight gains are observed in per-genre settings. This suggests that authors in this genre may share reflective thematic structures, while still preserving enough stylistic individuality to support model learning.

Society \& Reality remains volatile, with signs of intra-genre variability. It underperforms in zero-shot, especially in per-genre mode (F1 Macro: 0.3333 in Test2). After finetuning, performance improves in cross-genre mode in Test1 (F1 Weighted: 0.4711), but per-genre results remain low. The likely cause is stylistic heterogeneity within the genre, as diverse topics and rhetorical strategies may blur authorial boundaries. While finetuning improves robustness in broader contexts, it does not meaningfully enhance per-genre discrimination.

Finally, Landscape \& Journey presents an interesting reversal post-finetuning. In the zero-shot setting, it performs excellently in per-genre mode—achieving perfect scores in Test2 (F1 Macro: 1.0000) and a strong F1 Weighted score of 0.7579 in Test1. However, after finetuning, its per-genre performance drops sharply in Test1 (F1 Macro: 0.1111), while cross-genre scores remain relatively stable. This suggests that finetuning may have disrupted useful genre-specific patterns, possibly due to overfitting to inter-genre noise or a loss of token-level cues in smaller, more structured genres.

The model also has variations on performing in different modes (per-genre vs cross-genre).

In zero-shot, per-genre consistently outperforms cross-genre, reflecting the model’s reliance on shared genre structure for attribution. For example, in Test2 zero-shot, per-genre reached 0.8229 F1 Macro vs. 0.5296 for cross-genre.

However, after finetuning, this mode gap narrows or disappears. In Test2 finetuned, both modes yield identical scores across all metrics. This suggests that finetuning may equalize the model’s reliance on genre cues by reinforcing author-specific traits independent of genre. This is encouraging, as it implies that the model is no longer “shortcutting” via genre similarity, but truly learning attribution-relevant features.

Therefore, hypothesis 2 is strong supported.

\subsection{Error Analysis}
Our results across both zero-shot and finetuned settings point to key limitations in dataset design, model generalization, and evaluation structure.

A central issue arises in Test2 per-genre, where both in the zero-shot and finetuned settings, Folklore \& Tradition and Landscape \& Journey achieve perfect scores (1.0000) across all metrics. It shows possible structural limitations in our dataset, e.g. low pairwise diversity and potentially easier synthetic examples. We believe the root cause lies in the sparsity of our dataset, which is further amplified by token-level augmentation. This process may artificially inflate class separability by introducing predictable lexical artifacts, leading to overly optimistic scores that do not accurately reflect the model's true generalization ability.

We also observe a persistent divergence across F1 metrics, most notably in Test2. In the zero-shot setting, this divergence is subtle but clear—F1 Macro consistently lags behind F1 Micro and F1 Weighted, indicating imbalanced label distributions and suggesting that the model’s predictions are skewed toward a dominant class. To account for this, we adopt F1 Macro as the primary evaluation metric throughout our analysis, as it better reflects performance across both classes in the presence of imbalance. This imbalance becomes more pronounced in the finetuned setting, especially in Test2 per-genre, where recall is often perfect while precision and F1 are close to zero. This pattern points to a thresholding failure, where the model predicts nearly all instances as belonging to a single class (typically positive), resulting in a collapse in precision. By contrast, cross-genre performance remains relatively stable across both zero-shot and finetuned settings, reinforcing the notion that the model generalizes more effectively across genres than within them. This may reflect a broader challenge in capturing fine-grained intra-genre stylistic variation, particularly for unseen authors, and is likely exacerbated when the training data lacks sufficient diversity or depth in those stylistic dimensions.

Thus, while finetuning improves robustness and generalization for some genres in Test1, both zero-shot and finetuned results in Test2 expose shared weaknesses stemming from dataset design. These include label and author imbalance, synthetic augmentation artifacts, and limited within-genre variation.

\section{Conclusion and Future Work}

Our study investigates the effectiveness of fine-tuned and zero-shot models for Chinese lyrics authorship attribution, focusing on two hypotheses: (1) that a fine-tuned model outperforms a zero-shot approach, and (2) that model performance varies across genres.

Our findings strongly support Hypothesis 2—model performance is highly genre-dependent, with structured genres like Folklore \& Tradition consistently outperforming more abstract or emotionally diffuse genres such as Love \& Romance. This variation persists across both zero-shot and fine-tuned settings and highlights the importance of genre-sensitive evaluation in authorship tasks.

Hypothesis 1 is only partially supported. While fine-tuning improves robustness and generalization in Test1, particularly for real-world data and challenging genres, its benefits are less pronounced—or even obscured—in Test2. This inconsistency is largely due to design limitations in Test2, which is small, sparse, and relies on synthetic augmentation. These properties amplify issues such as label imbalance, shallow lexical cues, and overly homogeneous pairings, all of which can distort evaluation outcomes and mask the model's true generalization capabilities.

Our work create an initial benchmark and analysis. Looking ahead, future work should prioritize the development of larger, more diverse, and naturally constructed test sets, especially ones that better reflect real-world stylistic variation and authorial diversity. Reducing reliance on token-level augmentation and balancing author representation across genres will improve the reliability and interpretability of attribution models. Additionally, exploring domain-adaptive pretraining may further enhance the model’s ability to generalize across both seen and unseen authors.

\textbf{Ethical Considerations:} While our study contributes to understanding stylistic authorship in Chinese lyrics, it also raises ethical concerns. Authorship attribution models, if misused, could undermine creative anonymity, reinforce biases across cultural or gendered writing styles, or be leveraged inappropriately in legal or commercial settings. We emphasize that such models should not be used to make definitive claims about authorship without human verification. Moreover, respecting the privacy and intent of lyricists—particularly in culturally sensitive or politically expressive genres—remains paramount.

The full dataset and the Python scripts for data processing, train and test splitting, zero shot, and fine tune are available at our GitHub repository\footnote{\url{https://github.com/Lyxxx2003/s25-llm-project}}.

\bibliography{custom}

\clearpage
\appendix
\section{Statistics}
\renewcommand{\thesubsection}{A.\arabic{subsection}}

\subsection{Raw Lyrics Statistics}
\label{sec:appendixA1}

\begin{table}[H]
\centering
\caption{Song Length Statistics (First 1000 Songs)}
\label{tab:song-length-1000}
\small
\begin{tabular}{lrr}
\toprule
\textbf{Metric} & \textbf{Value} \\
\midrule
Median Song Length & 43.0 \\
Mean Song Length & 45.696 \\
Max Song Length & 119.0 \\
Min Song Length & 16.0 \\
Lower Bound & 10.5 \\
Upper Bound & 78.5 \\
Number of Outliers & 31.0 \\
\bottomrule
\end{tabular}
\end{table}

\begin{figure}[H]
\centering
\includegraphics[width=\columnwidth]{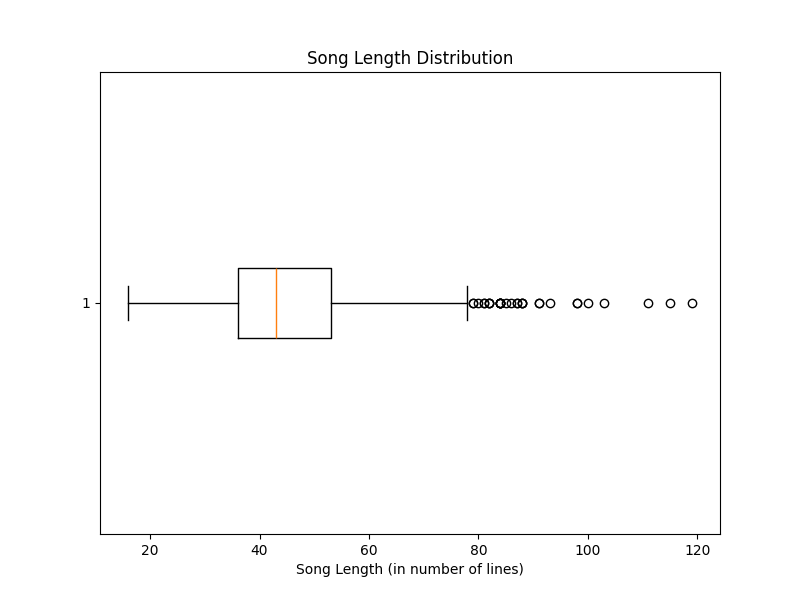}
\caption{Song length distribution for the first 1000 songs after data-processing.}
\label{fig:songlengthdistribution}
\end{figure}

\begin{table}[H]
\centering
\caption{Song Length Statistics (Last 55 Songs)}
\label{tab:song-length-55}
\small
\begin{tabular}{lrr}
\toprule
\textbf{Metric} & \textbf{Value} \\
\midrule
Median Song Length & 37.0 \\
Mean Song Length & 40.218 \\
Max Song Length & 78.0 \\
Min Song Length & 16.0 \\
\bottomrule
\end{tabular}
\end{table}

\subsection{Train Set Statistics}
\label{sec:appendixA2}

\begin{table}[H]
\centering
\caption{Train Set – Genre Breakdown}
\label{tab:train-genre}
\small
\begin{tabular}{lrrrrrrr}
\toprule
Genre & Authors & Pairs & Label 0 & Label 1 & L0\% & L1\% & \%Total \\
\midrule
Folklore \& Tradition & 16 & 44 & 19 & 25 & 43.18\% & 56.82\% & 4.56\% \\
Love \& Romance   & 90 & 505 & 241 & 264 & 47.72\% & 52.28\% & 52.33\% \\
Life \& Reflection & 80 & 373 & 191 & 182 & 51.21\% & 48.79\% & 38.65\% \\
Society \& Reality & 46 & 127 & 71 & 55 & 56.35\% & 43.65\% & 13.16\% \\
Landscape \& Journey & 45 & 139 & 67 & 72 & 48.20\% & 51.80\% & 14.40\% \\
\bottomrule
\end{tabular}
\end{table}

\vspace{10pt}

\begin{table}[H]
\centering
\caption{Train Set – Pair Type Breakdown}
\label{tab:train-pairtype}
\small
\begin{tabular}{lrrr}
\toprule
Type & Same-Author & Diff-Author & Total (\%) \\
\midrule
Per-genre & 303 (55.49\%) & 243 (44.51\%) & 546 (56.58\%) \\
Cross-genre & 168 (40.10\%) & 251 (59.90\%) & 419 (43.42\%) \\
\bottomrule
\end{tabular}
\end{table}

\clearpage
\subsection{Test1 Set Statistics}
\label{sec:appendixA3}

\begin{table}[H]
\centering
\caption{Test1 Set – Genre Breakdown}
\label{tab:test1-genre}
\small
\begin{tabular}{lrrrrrrr}
\toprule
Genre & Authors & Pairs & Label 0 & Label 1 & L0\% & L1\% & \%Total \\
\midrule
Folklore \& Tradition & 10 & 7 & 4 & 3 & 57.14\% & 42.86\% & 2.87\% \\
Love \& Romance   & 77 & 129 & 58 & 70 & 45.31\% & 54.69\% & 52.87\% \\
Life \& Reflection & 73 & 90 & 44 & 46 & 48.89\% & 51.11\% & 36.89\% \\
Society \& Reality & 35 & 31 & 16 & 15 & 51.61\% & 48.39\% & 12.70\% \\
Landscape \& Journey & 38 & 33 & 18 & 15 & 54.55\% & 45.45\% & 13.52\% \\
\bottomrule
\end{tabular}
\end{table}

\vspace{10pt}

\begin{table}[H]
\centering
\caption{Test1 Set – Pair Type Breakdown}
\label{tab:test1-pairtype}
\small
\begin{tabular}{lrrr}
\toprule
Type & Same-Author & Diff-Author & Total (\%) \\
\midrule
Per-genre & 74 (53.24\%) & 65 (46.76\%) & 139 (56.97\%) \\
Cross-genre & 48 (45.71\%) & 57 (54.29\%) & 105 (43.03\%) \\
\bottomrule
\end{tabular}
\end{table}

\subsection{Test2 Set Statistics}
\label{sec:appendixA4}

\begin{table}[H]
\centering
\caption{Test2 Set – Genre Breakdown}
\label{tab:test2-genre}
\small
\begin{tabular}{lrrrrrrr}
\toprule
Genre & Authors & Pairs & Label 0 & Label 1 & L0\% & L1\% & \%Total \\
\midrule
Folklore \& Tradition & 3 & 5 & 1 & 3 & 25.00\% & 75.00\% & 7.35\% \\
Love \& Romance & 13 & 12 & 9 & 2 & 81.82\% & 18.18\% & 17.65\% \\
Life \& Reflection & 13 & 24 & 14 & 10 & 58.33\% & 41.67\% & 35.29\% \\
Society \& Reality & 9 & 11 & 5 & 5 & 50.00\% & 50.00\% & 16.18\% \\
Landscape \& Journey & 9 & 16 & 6 & 4 & 60.00\% & 40.00\% & 23.53\% \\
\bottomrule
\end{tabular}
\end{table}

\vspace{10pt}

\begin{table}[H]
\centering
\caption{Test2 Set – Pair Type Breakdown}
\label{tab:test2-pairtype}
\small
\begin{tabular}{lrrr}
\toprule
Type & Same-Author & Diff-Author & Total (\%) \\
\midrule
Per-genre & 14 (41.18\%) & 20 (58.82\%) & 34 (50.00\%) \\
Cross-genre & 14 (41.18\%) & 20 (58.82\%) & 34 (50.00\%) \\
\bottomrule
\end{tabular}
\end{table}

\clearpage
\section{Prompts}
\renewcommand{\thesubsection}{B.\arabic{subsection}}
\subsection{Genre Generation}
\label{sec:genre-chinese-prompt}

The exact Chinese prompt used to query Deepseek-LLama-70B is shown below:

\begin{quote}
\begin{CJK}{UTF8}{gbsn}
给定以下的流派概念：\\
1. 爱与浪漫：浪漫、心碎、渴望、想念某人\\
2. 生活与反思：成长、遗憾、个人教训、沉思\\
3. 社会与现实：城市斗争、不平等、阶级、政治色彩\\
4. 风景与旅程：大自然、旅行、风景、漫游、孤独\\
5. 民俗与传统：传说、文化图标、地区故事、历史主题\\[0.5em]

请对以下歌词进行分类：\\
\{lyrics\}\\[0.5em]

如果模型认为多个流派几乎同样有可能是这首歌词的流派，可以生成多个流派。\\
但大多数情况下，应该只提供一个流派。输出应该遵循以下格式：\\[0.5em]

流派： [流派1]\\[0.5em]

其中，流派1是流派标签（例如，爱与浪漫，生活与反思）。\\
如果多个流派几乎同样有可能，则应将它们列在方括号内，并用逗号分隔，例如：\\
流派： [爱与浪漫, 生活与反思]。\\
但请注意，大多数情况下应该只有一个流派列在方括号内。
\end{CJK}
\end{quote}

\subsection{Zero-Shot Attribution}
\label{sec:chinese-prompt}

The exact Chinese prompt used to query Deepseek-LLama-70B is shown below:

\begin{quote}
\begin{CJK}{UTF8}{gbsn}
验证两段输入文本是否由同一位作者撰写。分析输入文本的写作风格，忽略主题和内容的差异。推理应基于语言特征，例如动词、标点符号、稀有词汇、词缀、幽默、讽刺、打字错误和拼写错误等。输出应遵循以下格式：0 或 1（0表示不同作者，1表示相同作者）。
\end{CJK}
\end{quote}

\section{Comparison Between Chinese And English Prompts in Genre Generation}
\renewcommand{\thesubsection}{C.\arabic{subsection}}
\label{sec:genre-prompt-comparison}

The following two examples provide the comparison of genre-generation results of Chinese and English prompts on two songs in the sample dataset. We conclude that the genres generated by the Chinese prompt in these two examples are more accurate than the ones generated by the English prompt.

\subsubsection*{Example 1:}

\begin{quote}
\begin{CJK}{UTF8}{gbsn}

\textbf{Title}: 望 \\
\textbf{Lyricist(s)}: 刘畅 \\
\textbf{Genre}: 生活与反思 \\
\textbf{Length}: 48 \\
\textbf{Lyrics}:

\begin{verbatim}
荆棘之上仰望的人
会看透风尘
在这善恶彼伏的城
我与你共存
那一束光透着锋芒
指引我穿越悲伤
那一次的遇见
命中注定是你
欲望将心磨去棱角
嘲笑温情从不胜仗
我偏要做这野火烧不尽的粮
把爱喂养
站在巨浪的中心 风是我思想
雨是催马还击的猎场
抬头遥望那些回不去的时光
眼前的火炬在激荡
天越来越黑 路越来越长
我知道人心不能摇晃
敢乱世为王 必掷血屠狼
才有希望
梦里有过渴望
醒来却容不得泪千行
曾经暗度陈仓
如今正面对抗
不接受 命运的盘问
爱恨教我以柔克刚
付出血的代价还要一颗心脏
寸寸偿
站在巨浪的中心 风是我思想
雨是催马还击的猎场
抬头遥望那些回不去的时光
眼前的火炬在激荡
天越来越黑 路越来越长
我知道人心不能摇晃
敢乱世为王 必掷血屠狼
才有希望
荆棘之上仰望的人
会看透风尘
在这善恶彼伏的城
我与你共存
穿过刀山 越过火海 把爱信仰
一颗真心有一片天堂
生如蜉蝣却要活得荡气回肠
相信前方就是故乡
我抬头仰望 这荆棘之上
大地的力量终将被释放
看万物生长 看日月沧桑
为你守望
\end{verbatim}

\end{CJK}
\end{quote}

\textbf{Chinese result:}
\begin{quote}
Raw API Response: \texttt{<think>} \\
\begin{CJK}{UTF8}{gbsn}
好，我现在要分析这首歌词的流派。首先，通读歌词，发现里面有很多关于奋斗、面对困难、坚持信念的内容。比如“站在巨浪的中心”、“敢乱世为王 必掷血屠狼”这些句子，显示出强烈的斗争精神和对抗困境的态度。 \\

接着，注意到歌词中多次提到“爱”和“信仰”，如“把爱喂养”、“穿过刀山 越过火海 把爱信仰”，这可能涉及到爱与浪漫的元素。但是，这里的“爱”更多是作为一种信念和力量，而不是单纯的浪漫情感。 \\

再看看是否有社会或现实的主题。歌词中提到“善恶彼伏的城”、“人心不能摇晃”，这可能涉及到社会中的道德和人性的考验，属于社会与现实的范畴。但整体来看，这些内容更多是为了支持歌曲的主线，即在困境中坚持和奋斗。 \\

还有关于反思和成长的内容，如“抬头遥望那些回不去的时光”、“曾经暗度陈仓 如今正面对抗”，这可能涉及到生活与反思的主题。但同样，这些内容服务于歌曲的核心主题，即在逆境中成长和坚持。 \\

综合来看，歌词的核心主题是面对困难、坚持信念、奋斗和成长，这些都属于生活与反思的流派。虽然涉及到爱和社会元素，但它们是为了支持主线，而不是主要焦点。因此，这首歌词应该归类为生活与反思。
\end{CJK}
\texttt{</think>} \\[0.5em]

\begin{CJK}{UTF8}{gbsn}
流派：[生活与反思]
\end{CJK}
\end{quote}

\textbf{English result:}
\begin{quote}
Raw API Response: \texttt{<think>} \\
Okay, so I need to classify the genre of these lyrics based on the given concepts. Let me read through the lyrics first and see what themes stand out. \\

The lyrics start with “\begin{CJK}{UTF8}{gbsn} 荆棘之上仰望的人 会看透风尘 \end{CJK}” which translates to something like “The one who looks up from the thorns can see through the world's dust.” That gives me a sense of struggle and resilience. The mention of “\begin{CJK}{UTF8}{gbsn} 善恶彼伏的城 \end{CJK}” or a city where good and evil are everywhere suggests a focus on society and its issues. \\

There's a lot about overcoming hardships, like “\begin{CJK}{UTF8}{gbsn} 穿过刀山 越过火海 把爱信仰 \end{CJK}” which means enduring severe trials. The line “\begin{CJK}{UTF8}{gbsn} 敢乱世为王 必掷血屠狼 \end{CJK}” implies a struggle for power and survival, which sounds political or societal. \\

The repetition of standing in the center of giant waves and the wind being thoughts, rain being a battlefield, shows a sense of turmoil and inner conflict. The mention of not accepting fate and fighting against it ties into themes of resistance and societal change. \\

There's also a strong sense of hope and perseverance, like “\begin{CJK}{UTF8}{gbsn} 相信前方就是故乡 \end{CJK}” or believing the front is home, indicating a journey or struggle towards a better place. The line about the earth's power being released and watching things grow suggests a connection to nature, but it's more about overcoming adversity than just nature itself. \\

Looking at the genre concepts, Society \& Reality deals with urban struggles, inequality, and political tones. The lyrics definitely touch on societal issues, the struggle between good and evil, and the fight against oppressive systems. While there are elements of personal growth and hope, the overarching theme seems to be about society and the reality of living in a challenging world. \\

I don't see much about romance or folklore here. It's more about the external struggles rather than internal personal reflections or love. So, I think Society \& Reality fits best. \\
\texttt{</think>} \\[0.5em]

Genres: [Society \& Reality]
\end{quote}

\subsubsection*{Example 2:}

\begin{quote}
\begin{CJK}{UTF8}{gbsn}

\textbf{Title}: 无虞 \\
\textbf{Lyricist(s)}: 刘畅 \\
\textbf{Genre}: [爱与浪漫, 生活与反思] \\
\textbf{Length}: 43 \\
\textbf{Lyrics}:

\begin{verbatim}
也许是雪落无声
也许是镜花无迹
看遍了岁月流转的场景
不及你 不及你
我注视一缕涟漪
好似无风亦无晴
只是爱越来得无声无息
心越点水化雨
你的眉宇有几分盛气
让我用尽一生去追寻
莫怪情话零星 闲了梦境
醒来更薄情
我这颗离情难耐的心
其实最藏不住心事
低头寻找的光景 道别后清晰
每一瞥都是想你
哪怕从此陌路相依
曾问知否那人心
耗尽最后的希冀
倘若留风不住便随风去
都无虞 都无虞
谁知哪一眼执迷
蒙尘过后仍清晰
若这一身傲骨不屈别离
往事皆可提
你的眉宇有几分盛气
让我用尽一生去追寻
莫怪情话零星 闲了梦境
醒来更薄情
我这颗离情难耐的心
其实最藏不住心事
低头寻找的光景 道别后清晰
每一瞥都是想你
哪怕从此陌路相依
我依然记得来时心情
才更难忘你每个踪影
那一年的自己仅凭天意
花光一生运气
原来我这平庸的一生
只为你回首不委屈
当一切归于平静
我也好看清月下是谁的身影
原来你早已注定
\end{verbatim}

\end{CJK}
\end{quote}

\textbf{Chinese result:}
\begin{quote}
Raw API Response: \texttt{<think>} \\
\begin{CJK}{UTF8}{gbsn}
好，我现在要分析这首歌词的流派。首先，通读歌词，发现里面有很多关于爱情的描述，比如“想你”、“追寻”、“心事”等等，这些都是典型的爱与浪漫的主题。接着，歌词中还提到了成长和回忆，比如“岁月流转的场景”、“道别后清晰”，这可能涉及到生活与反思的元素。再看看是否有其他流派的痕迹，比如社会与现实、风景与旅程或者民俗与传统，但歌词中并没有明显涉及这些方面的内容。所以，主要的流派应该是爱与浪漫，可能还有一点生活与反思的因素，但爱与浪漫更为突出。
\end{CJK}
\texttt{</think>} \\[0.5em]

\begin{CJK}{UTF8}{gbsn}
流派：[爱与浪漫, 生活与反思]
\end{CJK}
\end{quote}

\textbf{English result:}
\begin{quote}
Raw API Response: \texttt{<think>} \\
Okay, so I need to classify the genre of these lyrics based on the given concepts. Let me read through the lyrics first and see what they're about. \\

The lyrics start with “\begin{CJK}{UTF8}{gbsn} 也许是雪落无声，也许是镜花无迹 \end{CJK}” which translates to something like “Maybe the snow fell silently, maybe the flowers in the mirror left no trace.” This seems poetic and introspective, talking about silent and unnoticed things. Then it says “\begin{CJK}{UTF8}{gbsn} 看遍了岁月流转的场景，不及你，不及你 \end{CJK}” which means “I've seen all the changing scenes of the years, but none compare to you.” So there's a sense of longing and comparison, focusing on someone special. \\

Next lines: “\begin{CJK}{UTF8}{gbsn} 我注视一缕涟漪，好似无风亦无晴，只是爱越来得无声无息，心越点水化雨。 \end{CJK}” This translates to “I gaze at a ripple, as if there's no wind or clear sky, love comes silently and without a trace, my heart is like a gentle rain.” This is definitely about love and the quiet, persistent nature of it. \\

Then it mentions “\begin{CJK}{UTF8}{gbsn} 你的眉宇有几分盛气，让我用尽一生去追寻。 \end{CJK}” which is “Your eyebrows have a bit of pride, I'll spend my life pursuing you.” This shows dedication and pursuit in love, which is a strong romantic theme. \\

The lines “\begin{CJK}{UTF8}{gbsn} 莫怪情话零星，闲了梦境，醒来更薄情。 \end{CJK}” mean “Don't blame the scattered love words, idle dreams, waking up to even thinner love.” This seems to touch on heartbreak or the fleeting nature of love, which is part of the Love \& Romance genre. \\

Further on, “\begin{CJK}{UTF8}{gbsn} 我这颗离情难耐的心，其实最藏不住心事。 \end{CJK}” translates to “This heart that can't bear to part, actually can't hide its feelings.” Again, focusing on the heart and emotions related to love. \\

The repetition of “\begin{CJK}{UTF8}{gbsn} 你的眉宇有几分盛气，让我用尽一生去追寻 \end{CJK}” emphasizes the romantic pursuit. The mention of “\begin{CJK}{UTF8}{gbsn} 道别后清晰，每一瞥都是想你 \end{CJK}” which is “After saying goodbye, it's clear, every glance thinks of you,” shows longing and missing someone. \\

There's also a part about “\begin{CJK}{UTF8}{gbsn} 哪怕从此陌路相依，曾问知否那人心，耗尽最后的希冀。 \end{CJK}” meaning “Even if we become strangers relying on each other, I've asked if you know that person's heart, exhausting the last hope.” This indicates a sense of regret and longing, which fits into both Love \& Romance and maybe a bit of Life \& Reflection. \\

However, the overall theme seems to revolve around love, heartbreak, and longing. The focus is on the emotions towards a person, the pursuit, and the pain of separation. While there are elements of reflection, the primary theme is romantic. \\

So, considering the genres provided, Love \& Romance is the most fitting because the lyrics consistently talk about love, heartbreak, and missing someone. There's some reflection, but it's secondary to the romantic elements. \\
\texttt{</think>} \\[0.5em]

Genres: [Love \& Romance]
\end{quote}

\end{document}